\documentclass{article} % For LaTeX2e
\usepackage{nips15submit_e,times}
\usepackage{hyperref}
\usepackage{url}
\usepackage{bibhacks} % makes references header subsubsection level
\usepackage{natbib}
\usepackage[nolist]{acronym}
\usepackage{algorithm}
\usepackage{algorithmic}
\usepackage{graphicx}
\title{Fast Adaptive Weight Noise}

\author{
Justin Bayer \And Maximilian Karl \And Daniela Korhammer \\
Technische Universit\"at M\"unchen\\
\AND Patrick van der Smagt  \\
Technische Universit\"at M\"unchen\\
\textsl{and}
fortiss, TUM Associate Institute \\
\texttt{bayer.justin@googlemail.com, karlma@in.tum.de}\\
\texttt{korhammd@in.tum.de, smagt@brml.org} \\
}

\nipsfinalcopy % Uncomment for camera-ready version
\PassOptionsToPackage{fleqn}{amsmath}		% math environments and more by the AMS 
\usepackage{amsmath}
\usepackage{amssymb}
\usepackage{lscape,longtable}
\usepackage{todonotes}

\usepackage[makeroom]{cancel}

\newcommand{\eq}[1]{\begin{align*}#1\end{align*}}
\newcommand\numberthis{\addtocounter{equation}{1}\tag{\theequation}}

\newcommand{\bi}{\begin{itemize}}
\newcommand{\ei}{\end{itemize}}

\newcommand{\presupidx}[1]{{}^{#1}}

\newcommand*\kl[2]{\mathbb{KL}[#1||#2]}

\newcommand\diag{\operatorname{diag}}
\newcommand{\bA}{\mathbf{A}}
\newcommand{\ba}{\mathbf{a}}
\newcommand{\bb}{\mathbf{b}}
\newcommand{\bC}{\mathbf{C}}

\newcommand{\bmm}{\mathbf{m}}

\newcommand{\bV}{\mathbf{V}}
\newcommand{\bw}{\mathbf{w}}
\newcommand{\bW}{\mathbf{W}}
\newcommand{\bx}{\mathbf{x}}

\newcommand{\bu}{\mathbf{u}}
\newcommand{\bv}{\mathbf{v}}
\newcommand{\by}{\mathbf{y}}

\newcommand{\bz}{\mathbf{z}}

\newcommand{\beps}{\boldsymbol{\epsilon}}

\newcommand{\lossvi}{\mathcal{L}_{\text{vi}}}

\newcommand{\lossawn}{\mathcal{L}_{\text{awn}}}
\newcommand{\lossfawn}{\mathcal{L}_{\text{fawn}}}

\newcommand{\trainset}{\mathcal{D}_{\text{train}}}

\newcommand{\expc}[1]{\text{E}\left[#1\right]}
\newcommand{\vari}[1]{\text{V}[#1]}
\newcommand{\cova}[1]{\text{cov}[#1]}

\usepackage{cleveref}
\begin{document}

\maketitle

\begin{abstract}
Marginalising out uncertain quantities within the internal representations or parameters of neural networks is of central importance for a wide range of learning techniques, such as empirical, variational or full Bayesian methods.
We set out to generalise fast dropout \citep{wang2013fast} to cover a wider variety of noise processes in neural networks.
This leads to an efficient calculation of the marginal likelihood and predictive distribution which evades sampling and the consequential increase in training time due to highly variant gradient estimates.
This allows us to approximate variational Bayes for the parameters of feed-forward neural networks.
Inspired by the minimum description length principle, we also propose and experimentally verify the direct optimisation of the regularised predictive distribution.
The methods yield results competitive with previous neural network based approaches and Gaussian processes on a wide range of regression tasks.

\end{abstract}

\section{Introduction}
Deep learning methods have started to become practical for a wide range of tasks where very many labeled examples for supervised training are available, especially in the domains of sensory processing (e.g., vision or audio tasks).
Yet, methods which work well on data sets with few training cases in the context of regression of continuous quantities remain scarce.
Frequentist schemes such as weight decay or heuristics such as dropout have so far not been able to deliver significant improvements over methods not stemming from the connectionist paradigm, such as Gaussian processes or random forests; consequently, deep learning methods are generally not considered in fields where learning should be realised on small data sets.

We consider neural networks with parameters $\theta$ as weights and biases. If we treat the parameters $\theta$ not as points, but summarise our belief about them via a distribution $q(\theta)$, the data is explained by marginalising out that distribution, i.e.
\eq{
    p(\trainset) &= \int_\theta p(\trainset|\theta)q(\theta)d\theta  \numberthis \label{eq:marginal_likelihood}.
}
In the case of $p(\theta) = q(\theta)$, i.e. $q$ is a prior, this is commonly referenced to as the \emph{marginal likelihood}.
We will consider supervised data only, that is $\trainset = \{(\presupidx{i}\bx, \presupidx{i}\bz)\}_{i=1}^N$ where the functional relationship $\bx \rightarrow \bz$ is of interest.
In Bayesian learning, a prior $p(\theta)$ is used for $q$ to obtain a posterior via Bayes' rule
\eq{
    p(\theta|\trainset) &= {p(\theta) p(\trainset|\theta) \over p(\trainset)}, \numberthis \label{eq:posterior}
}
which can then be used to form a predictive distribution for unseen data points
\eq{
    p(\bz|\bx, \trainset) &= \int_\theta p(\bz|\bx, \theta) \, p(\theta|\trainset) \, d\theta.  \numberthis \label{eq:predictive}
}
In practice, Bayesian models are designed in a hierarchical way, where the prior is specified with the help of an additional hyperprior, i.e., $p(\theta) = \int_\eta p(\theta|\eta)p(\eta)$.

In all but the most trivial cases, Bayesian learning comes with several difficulties which require approximations.
For neural networks, not only will the posterior $p(\theta|\trainset)$ be highly multimodal due to symmetries in the weight space, but it will also be intractable to find the normalisation constant $p(\trainset) = \int p(\trainset|\theta)p(\theta)d\theta$, i.e., the marginal likelihood.

Due to this intractability for all but the simplest cases, neural network practitioners have to resort to approximation schemes such as sampling (e.g., via Markov chain Monte Carlo \cite{neal1993probabilistic}), variational inference \citep{hinton1993keeping}, combinations thereof \citep{graves2011practical}, or Gaussian approximations \citep{mackay1992practical}.

The contributions of this work will be as follows.
We will extend the idea of fast dropout \citep{wang2013fast} to the marginalisation of distributions over the weights of a neural network in \Cref{sec:varprop}, and introduce an efficient way to respect the correlations between outputs units in \Cref{sec:varprop_covar}.
This will be used to perform variational Bayes for the special case of a Gaussian likelihood function in \Cref{sec:fastvi}.
In \Cref{sec:fawn2} we will then propose a novel method to find a distribution over weights, namely the minimisation of the negative log-likelihood of the data plus a regularisation term.
The proposed methods will be verified experimentally in \Cref{sec:experiments}. 
We model the distributions over weights using diagonal Gaussian as well as Bernoulli distributions.

\subsection{Related Work}
The idea to treat weights in a neural network in a stochastic way, i.e.\ impose a distribution on them, goes back at least to \citet{buntine1991bayesian}.
Albeit dated, \citet{mackay1995probable} is an excellent survey article on probabilistically motivated approaches to neural networks, containing many concepts and ideas from the literature.
Using sampling-based techniques, \citet{graves2011practical} develops a practical algorithm based on \ac{VI}.
More recently, \citet{hernandez2015probabilistic} developed a method to treat units in neural networks in terms of their first two moments; they also develop a novel Bayesian learning algorithm for such scenarios. 
Most relevant to this section are the results from \citet{wang2013fast}---in fact, their work served as a starting point for this paper.
Even more recently \cite{blundell_weight_2015} used stochastic weights and an objective function using the variational free energy.
They used the reparametrisation trick from \cite{kingma2013auto} to backpropagate through the sampling process itself.
Most close to our work is the independently developed method by \cite{kingma2015variational}, who use fast dropout-like calculations to reduce the sampling effort and variance of the gradient estimators.

\section{Variance Propagation}
\label{sec:varprop}
\subsection{Propagation of Variance through a Transformation}
\label{sec:vp_linear_model}

We are using Variance Propagation to compute the effect of marginalising out the weight distribution.
This variance propagation is based on the works of \citet{wang2013fast}, where it was shown for the case of $\tilde\bx = \bx \cdot \bmm$ where $m_i \sim \mathcal{B}(d)$ follows a Bernoulli distribution with rate $d$.
Here, $\tilde\bx$ is the input to the model corrupted by ``dropout'' noise.

\citet{wang2013fast} provides a set of rules for propagation of mean and variance through a network. Rules for multiplication and addition are defined by elementary facts of probability \citep{grimmett1992probability}.

\subsubsection{Propagation of Variance through a Linear Transformation}

We have a linear transformation $a = \tilde\bw^T\tilde\bx + \tilde b$.
If $\tilde\bw, \tilde\bx$ and $\tilde b$ are independent of each other, have sufficiently many components and finite mean and variance, the central limit theorem \citep{grimmett1992probability} applies.
This makes $a$ distributed approximately according to a Gaussian, i.e. $a \sim \mathcal{N}(\expc{a}, \vari{a})$.
More specifically, consider a distribution $q(\tilde \theta)$ over the parameters of the model with $\tilde\theta = \{\tilde\bw, \tilde b\}$.

We obtain an approximation of the marginal likelihood (cf. \Cref{eq:marginal_likelihood}):
\eq{
    p(a|\bx) 
        &= \int_{\tilde\theta} q(\tilde\theta) p(a|\bx, \tilde\theta) d \tilde\theta \numberthis \label{eq:vp-marginal} \\
        &\approx \mathcal{N}(\expc{a}, \vari{a}).
}
All that is left to determine is then the expectation and variance of $a$.
Since both are sums and/or products of quantities with known expectation and variance, the calculations are given by
\eq{
    \expc{a} =& \expc{\tilde\bw}^T\expc{\tilde\bx} + \expc{\tilde b}, \label{eq:vp_expc_a} \numberthis \\
    \vari{a} 
        =& \vari{\tilde b} + \vari{\tilde \bw}^T\expc{\tilde \bx}^2 + \vari{\tilde \bx}^T\expc{\tilde \bw}^2 + \vari{\tilde \bx}^T\vari{\tilde \bw}, \label{eq:vp_vari_a} \numberthis
}
where we have assumed once again that all components of $\tilde \bx, \tilde b$ and $\tilde \bw$ are independent.

\subsubsection{Propagation of Variance through a Non-linear Function}

While the propagation through transfer functions is not in general tractable, the fact that the integral is one-dimensional allows for a wide range of approximations.
The most straightforward is the use of a table.
Other options include Monte Carlo integration and the unscented transform \citep{julier1997new}.
For the rectifier transfer and the logistic sigmoid function, a closed form and a very good approximation are available, respectively.
We present one of them here for the sake of completeness, but refer the interested reader to the corresponding paper by \citet{wang2013fast} for derivations.

In the case of the rectifier $f(a) = \text{max}(a, 0) = y$, we have:
\eq{
    r =& {\expc{a} \over \sqrt{\vari{a}}}, \\
    \expc{y} =& \Phi(r) \expc{a} + \phi(r) \sqrt{\vari{a}}, \\
    \vari{y} =& \expc{a} \sqrt{\vari{a}} \phi(r) + \bigl(\expc{a}^2 + \vari{a}\bigr) \Phi(r) - \expc{a}^2
}
where $\Phi(\xi)$ and $\phi(\xi)$ are the cumulative distribution function and probability density function of the standard Normal, respectively.

We want to stress the fact that propagating $\ba$ through the transfer function by integrating over each of its components $a_i$ separately will introduce the assumption that all elements of $\ba$ are statistically independent, which is certainly not completely justified.

\subsection{Variance Propagation for Deep Networks}
\label{sec:varprop_drnn}

In the previous section we described how to obtain the output expectation and variance of linear and non-linear transformations given the expectations and variances of its inputs.
Deep networks can be constructed by stacking many of these on top of each other. We apply these methods to multilayer perceptron networks with additional noise processes affecting the weights of the network.

It should be noted that all operations are differentiable and thus gradient-based optimisation can be used.
However, the equations are rather complex and use of an automatic differentiation tool such as Theano \citep{bergstra2010theano} is advisable.

\subsection{Noise Processes}

We now consider that the quantities $\tilde\bw, \tilde\bx, \tilde b$ are corrupted versions of the true underlying quantities $\bw, \bx, b$.
We will focus on $\tilde \bx$ first, while the discussion is equivalent for $\tilde \bw$ and $\tilde b$.
We define a noise process to be a probability distribution over possible corruptions given a clean input, i.e. $c(\tilde\bx|\bx)$.
If we can obtain $\expc{\tilde\bx}$ and $\vari{\tilde\bx}$ given $\expc{\bx}, \vari{\bx}$ and $c$, we can integrate $c$ seamlessly into the calculations.

Since we already gave the respective rules above, two obvious choices are additive and multiplicative noise.
Given a vector of independent noise variables $\beps$ with known expectation and covariance, let $\tilde\bx = \bx + \beps$, then
\eq{
     \expc{\tilde\bx} &= \expc{\bx} + \expc{\beps}, \\
     \vari{\tilde\bx} &= \vari{\bx} + \vari{\beps}.
}
Analogously, if $\tilde\bx = \bx \cdot \beps$, 
\eq{
     \expc{\tilde\bx} &= \expc{\bx} \cdot \expc{\beps}, \\
     \vari{\tilde\bx} &= \expc{\bx}^2 \vari{\beps} + \vari{\bx}\expc{\beps}^2 + \vari{\bx} \vari{\beps}.
}
Depending on the exact nature of $\beps$, several noise injecting regularisers can be approximated, such as Dropout \citep{hinton2012improving} (as done by \citet{wang2013fast}), DropConnect \citep{wan2013regularization} or Gaussian weight noise \citep{graves2013generating}.

\subsection{Soundness of the Approximation}
\label{sec:vp_soundness}
\citet{wang2013fast} verified experimentally that the central limit theorem holds for deep neural networks in certain cases.
This is, however, not possible in general and might fail in cases where inputs are low-dimensional or sparse. 
But is this at all important?
Considering that we are only interested in a function approximator, the exact interpretations of different quantities in the network are unimportant.
Loosely speaking, we do not care whether our model constitutes a good approximation of a corresponding real model, as long as the model works well enough for the task at hand, as indicated by an estimate of the generalisation error.

\section{Fast Adaptive Weight Noise}
\label{sec:fawn}

Adaptive weight noise is a practical method to perform \ac{VB} in neural networks \citep{graves2011practical}.
The method is based on the approach of \citet{hinton1993keeping}, who utilise the \ac{MDL} principle \citep{rissanen1985minimum,grunwald2007minimum} as an inductive bias.

As usual in the Bayesian setting, the parameters of the model under consideration are not found via point estimates, but represented as a distribution over the weight space.
Here, each parameter $\theta_i$ will be represented by a Gaussian, i.e. $q(\theta_i) = \mathcal{N}(\mu_i, \sigma^2_i)$.

If we are given a likelihood function and we consider $q$ as a variational approximation to the true posterior over the parameters having seen the data, the training criterion can be derived by means of \ac{VI}:
\eq{
    \lossvi :=& -\sum_i \int_{\theta} q(\theta) \log p(\presupidx{i}\bz|\presupidx{i}\bx, \theta) d\theta + \kl{q(\theta)}{p(\theta)} \numberthis \label{eq:loss-mdl} \\
             =& -\sum_i \expc{\log p(\presupidx{i}\bz|\presupidx{i}\bx, \theta)}_{\theta \sim q} + \kl{q(\theta)}{p(\theta)} \\
             \approx& -{1 \over S} \sum_i \sum_{s=1}^S \log p(\presupidx{i}\bz|\presupidx{i}\bx, \theta_s) + \kl{q(\theta)}{p(\theta)}, &~\theta_s \sim q(\theta) \numberthis \label{eq:loss-awn} \\
             =:& \lossawn,
}
where the outer sum is over the training samples.
The ``trick'' that \citet{hinton1993keeping} introduce is that the prior $p(\theta)$ is not set or further specified by a hyper-prior but instead \emph{learned as any other parameter in the model} and thus essentially set by data.
The contribution of \citet{graves2011practical} was then to approximate the expectation in \Cref{eq:loss-mdl} by Monte Carlo sampling with \Cref{eq:loss-awn}.

Here we use the previously introduced techniques to find a closed-form approximation to adaptive weight noise.
Consider a single layer with $\theta = \{\bw\}$, $y = f(\mathbf{x}^T\mathbf{w})$, where we have no dropout variables and the weights are Gaussian distributed with $\mathbf{w} \sim \mathcal{N}(\mu_{\mathbf{w}}, \sigma^2_{\mathbf{w}})$, with covariance diagonal and organised into a vector.
Again, we assume a Gaussian density for $a = \mathbf{x}^T\mathbf{w}$.
Using the rules from \Cref{sec:varprop}, we find that 
\eq{
\expc{a} &= \expc{\bx}^T\mu_{\bw}, \numberthis \label{eq:fawn-mean} \\
\vari{a} &= \vari{\bx}^T \mu_\bw^2 + \vari{\bx}^T\sigma_\bw^2 + (\expc{\bx}^2)^T\sigma_\bw^2. \numberthis \label{eq:fawn-var}
}
A perspective that we have not taken on so far is that this is a convolution of point predictions, each performed by a slightly different neural network with weights drawn from their respective distributions.
Consider a neural network $f(\bx, \theta)$ with $\theta = \{\theta_i\}$, where each $\theta_i$ is a Gaussian distributed random variable with mean $\mu_i$ and variance $\sigma^2_i$.
Let the network represent a distribution $p(\bz|\theta)$ for the random variable $\by$, which is the network's output.
The output of the network with marginalised weights will be approximated as such:
\eq{
    \int_{\theta} p(\bz|\bx, \theta) q(\theta)  d\theta \approx \mathcal{N}(\expc{\by}, \vari{\by}), \numberthis \label{eq:marg-weights}
}
where $q(\theta)$ depicts the joint over all weights and the moments of the Gaussian variable on the RHS are obtained as in \Cref{eq:fawn-mean,eq:fawn-var}.

\subsection{Output covariance}
\label{sec:varprop_covar}
While above we assumed all covariance matrices to be diagonal, we can easily and efficiently extend the last layer to explicitly model covariance in the output. 

Let $\bx$ be an input to the last layer and $\bW$ the weight matrix that maps this input to the output. Let $\bw_{*,o}$ and $\bw_{*,p}$ be two distinct columns of the weight matrix $\bW$ and $o=\bx^T \bw_{*,o}$ and $p = \bx^T \bw_{*,p}$ their respective outputs given $\bx$. In this model, we assume that $o$ and $p$ are not independent and need to extend the equation of variance propagation for addition
\eq{
	\vari{A + B} = \vari{A} + \vari{B} + 2\, \cova{A, B} \numberthis \label{eq:vari_add_cov}
}
for dependent outputs.

Plugging $o$ and $p$ into \cref{eq:vari_add_cov}, rearranging and using \cref{eq:vp_vari_a}, we can derive a simple formula for the covariance. (Note that since this assumes independence of $\bw_{*,o}$ and $\bw_{*,p}$, it does not hold for the diagonal entries of the covariance matrix.)
\eq{
	2 \cova{o, p} =& \vari{o + p} - \vari{o} - \vari{p}\\
				=& \vari{\bx^T \bw_{*,o} + \bb_o + \bx^T \bw_{*,p} + \bb_p} - \vari{\bx^T \bw_{*,o} + \bb_o} - \vari{\bx^T \bw_{*,p} + \bb_p}\\
				=& \vari{\bx^T (\bw_{*,o} + \bw_{*,p})} + \vari{\bb_o} + \vari{\bb_p} \\
				&- (\vari{\bx^T \bw_{*,o}} + \vari{\bb_o}) - (\vari{\bx^T \bw_{*,p}} + \vari{\bb_p})\\
				=& \vari{\bx^T (\bw_{*,o} + {\bw_{*,p}})} - \vari{\bx^T \bw_{*,o}} - \vari{\bx^T {\bw_{*,p}}}.
				 \numberthis \label{eq:cov}
}
By applying rules for variance propagation from \Cref{sec:vp_linear_model} and rearranging we arrive at:
\eq{
	\cova{o, p}  %=& \sum_i \vari{x_i} \expc{w_{io}}\expc{w_{ip}}\\
	          =& \vari{\bx}^T (\mu_{\bw_{*,o}}\circ \mu_{\bw_{*,p}}),
%	          =& \sigma_x^T (\mu_{\bw_{*,o}} \circ \mu_{\bw_{*,p}})
}
where $\circ$ denotes the Hadamard product.

We can show that the diagonal entries of the covariance matrix are computed in the same way as the variances of diagonal-covariance \ac{FAWN}.
%, that is
%\eq{ 
%	 \vari{o} &= \vari{\tilde b} + \vari{\tilde \bw_{*,o}}^T\expc{\tilde \bx}^2 + \cova{o, o} + \vari{\tilde \bx}^T\vari{\tilde \bw_{*,o}}.\\
%}
For the ``additional'' terms on the diagonal, we define
\eq{
	\bV = \diag(\sigma_{\bb}^2 + {\sigma_{\bw_{*,o}}^2}^T\expc{\bx}^2 + \vari{ \bx}^T \sigma_{\bw_{*,o}}^2)
}
and can then write the covariance matrix for full-covariance FAWN (Co-FAWN) in matrix notation:
\eq{
	\bC = \bV + \sum_i \bC_i = \bV + \sum_i \vari{x_i}  \mu_{\bw_{i, *}} \mu_{\bw_{i, *}}^T,
}
where $\bw_{i, *}$ is the $i$-th row of $\bW$.

\subsubsection{Computational efficiency}

For matrices, which are updated by adding the outer product of two vectors, the Sherman-Morrison formula,
\eq{
	(\bA + \bu \bv^T)^{-1} 
=	\bA^{-1} - \frac{\bA^{-1} \bu\bv^T \bA^{-1}}{1 + \bv^T \bA^{-1} \bu}, \numberthis \label{eq:sherman-morrison}
} 
presents a means of updating the inverse with an outer vector product. Similarly, the determinant of such a matrix can be updated using the matrix inversion lemma:
\eq{
	\det(\bA + \bu\bv^T) = (1 + \bv^T \bA^{-1} \bu) \det(\bA) \numberthis \label{eq:mil}
}
We define
\eq{
	\bA_{i+1} = \bA_i + \bu_i\bv_i^T
}
and can now recursively compute the determinant and inverse of $\bC$, which are needed to compute the loss, by setting $\bA_0 = \bV$, for which inversion and determinant computations are cheap,
\linebreak $\bu_i=\vari{x_i}  \expc{\bw_{i, *}}$ and $\bv_i = \expc{\bw_{i, *}}$ and repeatedly using \cref{eq:sherman-morrison,eq:mil} until we get the precision matrix $\bA_n^{-1} =\bC^{-1}$ and its determinant respectively.

The depth of the recursion corresponds to the number of hidden units in the last hidden layer $n$.

\subsection{Binary Weights}

In cases where memory and computational resources are limited, one can use Bernoulli distributed weights instead of Normal distributions.
This will half the amount of parameters needed. When using Bernoulli-distributed weights the same variance propagation rules  as for Gaussian distributed weights apply.
The only difference is in the mean and variance of the weight noise process:
\eq{
    \expc{\bw} &= (p-0.5)s, \numberthis \label{eq:exp_bern} \\
    \vari{\bw} &= p(1-p)s^2 \numberthis \label{eq:vari_bern}.
}
with $s$ as an additional weight scaler parameter and $p$ as the parameter defining a Bernoulli distribution.
This parameter $s$ helps the network to learn a richer set of functions since it would otherwise be limited to values between zero and one.
We compared the results against regular \ac{FAWN} as shown in \Cref{table:fawn} as FAWN-BERN.

\subsubsection{Justification by Sampling}

We compared the empirical distribution of outputs from the binary weights network with the variance propagation estimation.
Sampling from the output of a Bernoulli-distributed weights network is done by sampling weight matrices from the distribution of the weights $w^{\prime} \sim B(1,p)$ and scaling them with the parameters $s$ through $w = (w^{\prime} - 0.5)s$.
These sampled weight matrices are then used in a standard neural network to produce a sample from $p(\bz|\bx, \theta)$.
Histograms of these sampled outputs showed no significant deviation from the variance propagation approximation.

\subsection{Fast Variational Inference for Gaussian Likelihoods}
\label{sec:fastvi}

We will now use variance propagation to obtain an approximation to the first term of $\lossvi$ for the special case of a Gaussian likelihood.

Consider the first term of the RHS of \Cref{eq:loss-mdl} for the case that $\bz$ is assumed to be a univariate Gaussian. We will thus write $z$ for the targets and $y$ for the output of the network and leave out the dependency on $\theta$ for brevity.
Then,
\eq{
     & \expc{\log p(z|y)} \\
    =& \expc{\log \mathcal{N}(z|y, \sigma^2)} \\
    =& \expc{{-(z -y)^2 \over 2 \sigma^2} - \log \sqrt{2 \pi} \sigma} \\
    =& {-\expc{(z -y)^2} \over 2 \sigma^2} - \log \sqrt{2 \pi} \sigma \\
    =& -{\vari{y} \over 2\sigma^2} 
       -{(z - \expc{y})^2 \over 2\sigma^2}
       -\log\sqrt{2 \pi}\sigma  \\
    =& \log \mathcal{N}(\sqrt{\vari{y}}|0, \sigma^2)
       + \log \mathcal{N}(z|\expc{y}, \sigma^2) \\
       &+ \log \sqrt{2 \pi} \sigma.
}
where we have made use of the identity $\vari{y} = \expc{y^2} - \expc{y}^2$.
The last line offers a partially probabilistic interpretation of this specific instance of variational inference.
It puts a zero-centred prior on the square root of the output's variance and on the error, sharing the same (prior) variance---which is itself encouraged to be large.
The last term can be seen as a measure against the variance collapsing to zero, which would lead to large likelihoods on the training set.
We refer to this method as FAWN-VI.

\subsection{Optimisation of the predictive distribution with regularisation}
\label{sec:fawn2}

Since we now have an efficient approximation of the predictive distribution (cf.\ \Cref{eq:predictive} and \Cref{eq:marginal_likelihood}), an obvious next step is to directly optimise it with respect to the parameter distributions $q(\theta)$.
This will essentially lead to a maximum likelihood approach and thus inherit its tendency to overfit the training data.
Accounting for that is possible by a fully Bayesian treatment, which means to impose a hyperprior on $q(\theta)$ and integrate it out.

Here we shall follow a different route, which is to make use of a regulariser, namely the KL-divergence between $q(\theta)$ and a prior $p(\theta)$:
\eq{
    \lossfawn := -\sum_i \log \int_{\theta} q(\theta) p(\presupidx{i}\bz|\presupidx{i}\bx, \theta) d\theta + \kl{q(\theta)}{p(\theta)},
}
where the sum runs over the training samples $\trainset = \{(\presupidx{i}\bx, \presupidx{i}\bz)\}_{i=1}^N$.
We refer to this method as FAWN-ROPD.

\section{Experiments}
\label{sec:experiments}
We evalutated FAWN-VI and FAWN-ROPD from \Cref{sec:fastvi,sec:fawn2} respectively on a range of static regression tasks using \acp{FFN}. 
We are interested in finding not only a point prediction but a whole predictive distribution.
These tasks are typically not where neural networks excel and practicioners resort to \acp{GP} instead, which is why we compare to those.

To this end we used a global univariate Gaussian for the prior and a Gaussian as a variational approximation for each of the parameters:
\eq{
    p(\theta) =& \prod_i \mathcal{N}(\theta_i|\tilde\mu, \tilde\sigma^2), \\
    q(\theta) =& \prod_i \mathcal{N}(\theta_i|\dot\mu_i, \dot\sigma_i^2).
}
The KL-divergence is then given by\footnote{Obtained with the help of the Q\&A community ``crossvalidated'' at \url{http://stats.stackexchange.com/questions/7440/kl-divergence-between-two-univariate-gaussians}.}:
\eq{
    \kl{q(\theta)}{p(\theta)} = \sum_i \log \frac{\tilde\sigma}{\dot\sigma_i} + \frac{\dot\sigma_i^2 + (\dot\mu_i - \tilde\mu)^2}{2 \tilde\sigma^2} - \frac{1}{2}.
}

Additionally, we chose a Gaussian likelihood where we assumed that 
\eq{
    z_i = y_i + \epsilon_i, \epsilon_i \sim \mathcal{N}(0, \hat \sigma_i^2),
}
which resembles a Gaussian distributed measurement error with variance $\hat\sigma_i$ for output dimension $i$.
We integrate the $\hat\sigma_i$ into the set of parameters and optimise it jointly with all other parameters.

All experiments were performed using a similar protocol to the one used in \citet{hernandez2015probabilistic}: we used single-layer networks with 50 hidden units using the rectifier transfer function. 
We report the negative log likelihood of the data with means and standard deviations coming from ten different random splits into 90\% training and 10\% testing data.
The parameters of neural networks using \ac{FAWN} were drawn from a zero-centred Gaussian with standard deviation 0.2.

Training was performed using Adam \citep{kingma2014adam} with a step rate of $\alpha=0.001$ until convergence of the \emph{training loss}.
No separate validation set was used.
Gradients were estimated using 128 samples in a single mini batch. 

The results for \acp{GP} were obtained using a the sum of a linear and a squared exponential kernel using automatic relevance determination.
Three random restarts were performed.
We used GPy \citep{gpy2014} for the experiments.

The results are summarised in \Cref{table:fawn}.
The proposed methods place themselves well among alternative approaches, where FAWN-ROPD is better than FAWN-VI in all cases.

\begin{table*}[t]
\begin{center}
\caption[ ]{
    Results for \ac{FAWN}.
    Results for probabilistic backpropagation (PBP) and adaptive weight noise (VI) taken from \cite{hernandez2015probabilistic}. 
    Results for \acp{GP} obtained via GPy \citep{gpy2014}, where no results for the slightly bigger data sets (more than 1500 samples) were obtained due to the increased run time.
    Best results shown in bold.
}
\small
\setlength{\tabcolsep}{.15cm}
\renewcommand{\arraystretch}{1.2}
\begin{tabular}[b]{l|rrr|rrr}

    \label{table:fawn}
%                & \centering{N} & D & VI & PBP & GP & FAWN-VI & FAWN-LSVI \\
                & \multicolumn{1}{c}{VI} & \multicolumn{1}{c}{PBP} & \multicolumn{1}{c|}{GP} & \multicolumn{1}{c}{FAWN-VI} & \multicolumn{1}{c}{FAWN-ROPD} & \multicolumn{1}{c}{FAWN-BERN} \\
    \hline
    Boston      & 2.903$\pm$0.071            & \textbf{2.550}$\pm$\textbf{0.089}  & 2.631$\pm$0.289           & 3.005$\pm$0.273 & \textbf{2.559$\pm$0.161} & 2.685$\pm$0.196\\
    Concrete    & 3.391$\pm$0.017            & 3.136$\pm$0.021           & \textbf{2.893$\pm$0.095} & 3.183$\pm$0.077 & 3.107$\pm$0.134 & 3.310$\pm$0.109\\
    Energy      & 2.391$\pm$0.029            & 1.982$\pm$0.027           & \textbf{0.711$\pm$1.477}  & 1.762$\pm$0.655 & 1.369$\pm$0.842 & 2.095$\pm$0.077\\
    Kin8Nm      & 0.897$\pm$0.010            & -0.964$\pm$0.007          & \multicolumn1{c|}{--}                       & -1.006$\pm$0.027 & \textbf{-1.211$\pm$0.032} & -0.601$\pm$0.021\\
    Naval       & -3.734$\pm$0.116           & -3.653$\pm$0.004          & \multicolumn1{c|}{--}                       & -6.751$\pm$0.118& \textbf{-6.837$\pm$0.131} & -3.608$\pm$0.066\\
    Power Plant & 2.890$\pm$0.010            & 2.838$\pm$0.008           & \multicolumn1{c|}{--}                       & 2.849$\pm$0.042 & \textbf{2.819$\pm$0.029} & 2.859$\pm$0.031\\
    Protein     & 2.992$\pm$0.006            & 2.974$\pm$0.002           & \multicolumn1{c|}{--}                       & 2.973$\pm$0.022 & \textbf{2.882$\pm$0.068} & 3.005$\pm$0.013\\
    Wine        & 0.980$\pm$0.013            & 0.966$\pm$0.014           & \multicolumn1{c|}{--}                       & 0.943$\pm$0.037 & \textbf{0.908$\pm$0.078} & 0.934$\pm$0.085\\
    Yacht       & 3.439$\pm$0.163            & 1.483$\pm$0.018           & 0.615$\pm$0.756         & 1.448$\pm$0.393 & \textbf{0.336$\pm$0.271} & 3.201$\pm$0.191\\
    Year        & 3.622$\pm$~N/A~            & 3.603$\pm$~N/A~           & \multicolumn1{c|}{--}                       & 3.807$\pm$~N/A~ & \textbf{3.472$\pm$~N/A~} & \multicolumn1{c}{\,\,\,--}
    \end{tabular}
\end{center}
\end{table*}
\begin{table*}[t]
\begin{minipage}[t]{0.65\textwidth}

\begin{center}
\caption[ ]{
    Results for \ac{FAWN} and Co-FAWN used on multi-output datasets. For the Jura dataset we train on the location coordinates only and predict the local concentrations of the six different elements.
   % Results for probabilistic backpropagation (PBP) and adaptive weight noise (VI) taken from \cite{hernandez2015probabilistic}. 
    %Results for \acp{GP} obtained via GPy \citep{gpy2014}, where no results for the slightly bigger data sets (more than 1500 samples) were obtained due to the increased run time.
    Best results shown in bold.
}
\small
\setlength{\tabcolsep}{.15cm}
\renewcommand{\arraystretch}{1.2}
\begin{tabular}[b]{l|rrr|rr}

    \label{table:cofawn}
%                & \centering{N} & D & VI & PBP & GP & FAWN-VI & FAWN-LSVI \\
                & \multicolumn{1}{c}{N} & \multicolumn{1}{c}{D}& \multicolumn{1}{c|}{out} & \multicolumn{1}{c}{FAWN-ROPD} & \multicolumn{1}{c}{Co-FAWN}  \\
    \hline
    Energy      & 768     & 8  & 2 & 2.1218$\pm$0.7024           & \textbf{2.1063$\pm$0.8357} \\
    Naval       & 11'934  & 16 & 2 & -14.9868$\pm$0.7368         & \textbf{-15.1074$\pm$0.3656} \\
    Sarcos      & 48'933  & 21 & 7 & -4.4185$\pm$~N/A~~         & \textbf{-5.1867$\pm$~N/A~~~} \\
    Jura 	& 358     & 2  & 7 & 11.1407$\pm$~N/A~~          & \textbf{8.6396$\pm$~N/A~~~} \\
\end{tabular}
\end{center}
\end{minipage}
\hfill{}
\begin{minipage}[t]{0.3\textwidth}
\begin{center}
\caption[ ]{
    Size of Datasets
}
\small
\setlength{\tabcolsep}{.15cm}
\begin{tabular}[b]{l|rr}
    \label{table:datasetsize}
                & N & D \\
    \hline
    Boston      & 506     & 13 \\
    Concrete    & 1030    & 8  \\
    Energy      & 768     & 8  \\
    Kin8Nm      & 8192    & 8  \\
    Naval       & 11'934  & 16 \\
    Power Plant & 9568    & 4  \\
    Protein     & 45'730  & 9  \\
    Wine        & 1599    & 11 \\
    Yacht       & 308     & 6  \\
    Year        & 515'345 & 90
\end{tabular}
\end{center}
\end{minipage}
\end{table*}

\section{Conclusion and Future Work}
We have proposed a method to approximate the marginal likelihood of a distribution over neural network weights up to its mean and variance.
This enabled us to derive a deterministic approximation of variational Bayes for Gaussian likelihoods and propose a novel, less subjective flavour of variational inference, FAWN-ROPD.
The experimental results show that FAWN-ROPD obtains competitive performance over a wide range of regression tasks.
These tasks include ones with very little samples (order of a few hundred) as well as many samples (several thousands) and range from domains such as robotics, predictive maintenance, computational biology and others.

The method requires further evaluation: we will experimentally investigate more common deep-learning architectures such as recurrent neural networks and deep multilayer perceptrons.
Further, the suitability of FAWN for tasks where model uncertainty in the predictions is of interest, such as active learning or reinforcement learning, needs to be tested.
On the theoretical side, the exact relationship of FAWN-ROPD to reference priors remains unclear and a theoretically founded motivation for FAWN-ROPD is an important next step.

\begin{acronym}[Bash]
 \acro{FAWN}{Fast Adaptive Weight Noise}
 \acro{VB}{Variational Bayes}
 \acro{MDL}{Minimum Description Length}
 \acro{VI}{Variational Inference}
 \acro{FFN}{Feed-Forward Neural Network}
 \acro{GP}{Gaussian Processes}
\end{acronym}

%References follow the acknowledgments. Use unnumbered third level heading for
%the references. Any choice of citation style is acceptable as long as you are
%consistent. It is permissible to reduce the font size to `small' (9-point) 
%when listing the references. {\bf Remember that this year you can use
%a ninth page as long as it contains \emph{only} cited references.}
\clearpage
\bibliography{bibliography}

\begin{thebibliography}{21}
\providecommand{\natexlab}[1]{#1}
\providecommand{\url}[1]{\texttt{#1}}
\expandafter\ifx\csname urlstyle\endcsname\relax
  \providecommand{\doi}[1]{doi: #1}\else
  \providecommand{\doi}{doi: \begingroup \urlstyle{rm}\Url}\fi

\bibitem[Bergstra et~al.(2010)Bergstra, Breuleux, Bastien, Lamblin, Pascanu,
  Desjardins, Turian, Warde-Farley, and Bengio]{bergstra2010theano}
Bergstra, James, Breuleux, Olivier, Bastien, Fr{\'{e}}d{\'{e}}ric, Lamblin,
  Pascal, Pascanu, Razvan, Desjardins, Guillaume, Turian, Joseph, Warde-Farley,
  David, and Bengio, Yoshua.
\newblock Theano: a {CPU} and {GPU} math expression compiler.
\newblock In \emph{Proceedings of the Python for Scientific Computing
  Conference ({SciPy})}, June 2010.
\newblock URL
  \url{http://www.iro.umontreal.ca/~lisa/pointeurs/theano_scipy2010.pdf}.
\newblock Oral Presentation.

\bibitem[Blundell et~al.(2015)Blundell, Cornebise, Kavukcuoglu, and
  Wierstra]{blundell_weight_2015}
Blundell, Charles, Cornebise, Julien, Kavukcuoglu, Koray, and Wierstra, Daan.
\newblock Weight uncertainty in neural networks.
\newblock \emph{{arXiv}:1505.05424 [cs, stat]}, 2015.

\bibitem[Buntine \& Weigend(1991)Buntine and Weigend]{buntine1991bayesian}
Buntine, Wray~L and Weigend, Andreas~S.
\newblock Bayesian back-propagation.
\newblock \emph{Complex systems}, 5\penalty0 (6):\penalty0 603--643, 1991.

\bibitem[{GPy~authors}(2012--2014)]{gpy2014}
{GPy~authors}, The.
\newblock {GPy}: A {G}aussian process framework in python.
\newblock \url{http://github.com/SheffieldML/GPy}, 2012--2014.

\bibitem[Graves(2011)]{graves2011practical}
Graves, Alex.
\newblock Practical variational inference for neural networks.
\newblock In \emph{Advances in Neural Information Processing Systems}, pp.\
  2348--2356, 2011.

\bibitem[Graves(2013)]{graves2013generating}
Graves, Alex.
\newblock Generating sequences with recurrent neural networks.
\newblock \emph{arXiv preprint arXiv:1308.0850}, 2013.

\bibitem[Grimmett \& Stirzaker(1992)Grimmett and
  Stirzaker]{grimmett1992probability}
Grimmett, Geoffrey and Stirzaker, David.
\newblock \emph{Probability and random processes}, volume~2.
\newblock Oxford Univ Press, 1992.

\bibitem[Gr{\"u}nwald(2007)]{grunwald2007minimum}
Gr{\"u}nwald, Peter~D.
\newblock \emph{The minimum description length principle}.
\newblock MIT press, 2007.

\bibitem[Hern{\'a}ndez-Lobato \& Adams(2015)Hern{\'a}ndez-Lobato and
  Adams]{hernandez2015probabilistic}
Hern{\'a}ndez-Lobato, Jos{\'e}~Miguel and Adams, Ryan~P.
\newblock Probabilistic backpropagation for scalable learning of bayesian
  neural networks.
\newblock \emph{arXiv preprint arXiv:1502.05336}, 2015.

\bibitem[Hinton \& Van~Camp(1993)Hinton and Van~Camp]{hinton1993keeping}
Hinton, Geoffrey~E and Van~Camp, Drew.
\newblock Keeping the neural networks simple by minimizing the description
  length of the weights.
\newblock In \emph{Proceedings of the sixth annual conference on Computational
  learning theory}, pp.\  5--13. ACM, 1993.

\bibitem[Hinton et~al.(2012)Hinton, Srivastava, Krizhevsky, Sutskever, and
  Salakhutdinov]{hinton2012improving}
Hinton, Geoffrey~E, Srivastava, Nitish, Krizhevsky, Alex, Sutskever, Ilya, and
  Salakhutdinov, Ruslan~R.
\newblock Improving neural networks by preventing co-adaptation of feature
  detectors.
\newblock \emph{arXiv preprint arXiv:1207.0580}, 2012.

\bibitem[Julier \& Uhlmann(1997)Julier and Uhlmann]{julier1997new}
Julier, Simon~J and Uhlmann, Jeffrey~K.
\newblock New extension of the kalman filter to nonlinear systems.
\newblock In \emph{AeroSense'97}, pp.\  182--193. International Society for
  Optics and Photonics, 1997.

\bibitem[Kingma \& Ba(2014)Kingma and Ba]{kingma2014adam}
Kingma, Diederik and Ba, Jimmy.
\newblock Adam: A method for stochastic optimization.
\newblock \emph{arXiv preprint arXiv:1412.6980}, 2014.

\bibitem[Kingma \& Welling(2013)Kingma and Welling]{kingma2013auto}
Kingma, Diederik~P and Welling, Max.
\newblock Auto-encoding variational bayes.
\newblock \emph{arXiv preprint arXiv:1312.6114}, 2013.

\bibitem[Kingma et~al.(2015)Kingma, Salimans, and
  Welling]{kingma2015variational}
Kingma, Diederik~P, Salimans, Tim, and Welling, Max.
\newblock Variational dropout and the local reparameterization trick.
\newblock \emph{arXiv preprint arXiv:1506.02557}, 2015.

\bibitem[MacKay(1992)]{mackay1992practical}
MacKay, David~JC.
\newblock A practical bayesian framework for backpropagation networks.
\newblock \emph{Neural computation}, 4\penalty0 (3):\penalty0 448--472, 1992.

\bibitem[MacKay(1995)]{mackay1995probable}
MacKay, David~JC.
\newblock Probable networks and plausible predictions-a review of practical
  bayesian methods for supervised neural networks.
\newblock \emph{Network: Computation in Neural Systems}, 6\penalty0
  (3):\penalty0 469--505, 1995.

\bibitem[Neal(1993)]{neal1993probabilistic}
Neal, Radford~M.
\newblock Probabilistic inference using markov chain monte carlo methods.
\newblock Technical report, Department of Computer Science, University of
  Toronto Toronto, CA, 1993.

\bibitem[Rissanen(1985)]{rissanen1985minimum}
Rissanen, Jorma.
\newblock Minimum-description-length principle.
\newblock \emph{Encyclopedia of statistical sciences}, 1985.

\bibitem[Wan et~al.(2013)Wan, Zeiler, Zhang, Cun, and
  Fergus]{wan2013regularization}
Wan, Li, Zeiler, Matthew, Zhang, Sixin, Cun, Yann~L, and Fergus, Rob.
\newblock Regularization of neural networks using dropconnect.
\newblock In \emph{Proceedings of the 30th International Conference on Machine
  Learning (ICML-13)}, pp.\  1058--1066, 2013.

\bibitem[Wang \& Manning(2013)Wang and Manning]{wang2013fast}
Wang, Sida and Manning, Christopher.
\newblock Fast dropout training.
\newblock In \emph{Proceedings of the 30th International Conference on Machine
  Learning (ICML-13)}, pp.\  118--126, 2013.

\end{thebibliography}


\begin{thebibliography}{18}
\providecommand{\natexlab}[1]{#1}
\providecommand{\url}[1]{\texttt{#1}}
\expandafter\ifx\csname urlstyle\endcsname\relax
  \providecommand{\doi}[1]{doi: #1}\else
  \providecommand{\doi}{doi: \begingroup \urlstyle{rm}\Url}\fi

\bibitem[authors(2012--2014)]{gpy2014}
authors, The~GPy.
\newblock {GPy}: A gaussian process framework in python.
\newblock \url{http://github.com/SheffieldML/GPy}, 2012--2014.

\bibitem[Bergstra et~al.(2010)Bergstra, Breuleux, Bastien, Lamblin, Pascanu,
  Desjardins, Turian, Warde-Farley, and Bengio]{bergstra2010theano}
Bergstra, James, Breuleux, Olivier, Bastien, Fr{\'{e}}d{\'{e}}ric, Lamblin,
  Pascal, Pascanu, Razvan, Desjardins, Guillaume, Turian, Joseph, Warde-Farley,
  David, and Bengio, Yoshua.
\newblock Theano: a {CPU} and {GPU} math expression compiler.
\newblock In \emph{Proceedings of the Python for Scientific Computing
  Conference ({SciPy})}, June 2010.
\newblock URL
  \url{http://www.iro.umontreal.ca/~lisa/pointeurs/theano_scipy2010.pdf}.
\newblock Oral Presentation.

\bibitem[Bernardo(1979)]{bernardo1979reference}
Bernardo, Jose~M.
\newblock Reference posterior distributions for bayesian inference.
\newblock \emph{Journal of the Royal Statistical Society. Series B
  (Methodological)}, pp.\  113--147, 1979.

\bibitem[Buntine \& Weigend(1991)Buntine and Weigend]{buntine1991bayesian}
Buntine, Wray~L and Weigend, Andreas~S.
\newblock Bayesian back-propagation.
\newblock \emph{Complex systems}, 5\penalty0 (6):\penalty0 603--643, 1991.

\bibitem[Graves(2011)]{graves2011practical}
Graves, Alex.
\newblock Practical variational inference for neural networks.
\newblock In \emph{Advances in Neural Information Processing Systems}, pp.\
  2348--2356, 2011.

\bibitem[Graves(2013)]{graves2013generating}
Graves, Alex.
\newblock Generating sequences with recurrent neural networks.
\newblock \emph{arXiv preprint arXiv:1308.0850}, 2013.

\bibitem[Grimmett \& Stirzaker(1992)Grimmett and
  Stirzaker]{grimmett1992probability}
Grimmett, Geoffrey and Stirzaker, David.
\newblock \emph{Probability and random processes}, volume~2.
\newblock Oxford Univ Press, 1992.

\bibitem[Gr{\"u}nwald(2007)]{grunwald2007minimum}
Gr{\"u}nwald, Peter~D.
\newblock \emph{The minimum description length principle}.
\newblock MIT press, 2007.

\bibitem[Hern{\'a}ndez-Lobato \& Adams(2015)Hern{\'a}ndez-Lobato and
  Adams]{hernandez2015probabilistic}
Hern{\'a}ndez-Lobato, Jos{\'e}~Miguel and Adams, Ryan~P.
\newblock Probabilistic backpropagation for scalable learning of bayesian
  neural networks.
\newblock \emph{arXiv preprint arXiv:1502.05336}, 2015.

\bibitem[Hinton \& Van~Camp(1993)Hinton and Van~Camp]{hinton1993keeping}
Hinton, Geoffrey~E and Van~Camp, Drew.
\newblock Keeping the neural networks simple by minimizing the description
  length of the weights.
\newblock In \emph{Proceedings of the sixth annual conference on Computational
  learning theory}, pp.\  5--13. ACM, 1993.

\bibitem[Hinton et~al.(2012)Hinton, Srivastava, Krizhevsky, Sutskever, and
  Salakhutdinov]{hinton2012improving}
Hinton, Geoffrey~E, Srivastava, Nitish, Krizhevsky, Alex, Sutskever, Ilya, and
  Salakhutdinov, Ruslan~R.
\newblock Improving neural networks by preventing co-adaptation of feature
  detectors.
\newblock \emph{arXiv preprint arXiv:1207.0580}, 2012.

\bibitem[Julier \& Uhlmann(1997)Julier and Uhlmann]{julier1997new}
Julier, Simon~J and Uhlmann, Jeffrey~K.
\newblock New extension of the kalman filter to nonlinear systems.
\newblock In \emph{AeroSense'97}, pp.\  182--193. International Society for
  Optics and Photonics, 1997.

\bibitem[Kingma \& Ba(2014)Kingma and Ba]{kingma2014adam}
Kingma, Diederik and Ba, Jimmy.
\newblock Adam: A method for stochastic optimization.
\newblock \emph{arXiv preprint arXiv:1412.6980}, 2014.

\bibitem[MacKay(1995)]{mackay1995probable}
MacKay, David~JC.
\newblock Probable networks and plausible predictions-a review of practical
  bayesian methods for supervised neural networks.
\newblock \emph{Network: Computation in Neural Systems}, 6\penalty0
  (3):\penalty0 469--505, 1995.

\bibitem[Neal(1993)]{neal1993probabilistic}
Neal, Radford~M.
\newblock Probabilistic inference using markov chain monte carlo methods.
\newblock 1993.

\bibitem[Rissanen(1985)]{rissanen1985minimum}
Rissanen, Jorma.
\newblock Minimum-description-length principle.
\newblock \emph{Encyclopedia of statistical sciences}, 1985.

\bibitem[Wan et~al.(2013)Wan, Zeiler, Zhang, Cun, and
  Fergus]{wan2013regularization}
Wan, Li, Zeiler, Matthew, Zhang, Sixin, Cun, Yann~L, and Fergus, Rob.
\newblock Regularization of neural networks using dropconnect.
\newblock In \emph{Proceedings of the 30th International Conference on Machine
  Learning (ICML-13)}, pp.\  1058--1066, 2013.

\bibitem[Wang \& Manning(2013)Wang and Manning]{wang2013fast}
Wang, Sida and Manning, Christopher.
\newblock Fast dropout training.
\newblock In \emph{Proceedings of the 30th International Conference on Machine
  Learning (ICML-13)}, pp.\  118--126, 2013.

\end{thebibliography}
\bibliographystyle{icml2015}

\end{document}